\documentclass[twocolumn]{article}
\usepackage[utf8]{inputenc}
\usepackage[T1]{fontenc}
\usepackage{amsmath,amssymb}
\usepackage{graphicx}
\usepackage{natbib}
\usepackage{hyperref}
\usepackage{url}
\usepackage{booktabs}
\usepackage{float}
\usepackage[margin=1in]{geometry}

\setcitestyle{authoryear,open={(},close={)}}

\newcommand{\citeapos}[1]{\citeauthor{#1}'s (\citeyear{#1})}

\title{LLMs as Signal Detectors: Sensitivity, Bias, and the Temperature--Criterion Analogy}

\author{Jon-Paul Cacioli\\
Independent Researcher, Melbourne, Australia\\
\texttt{synthium@hotmail.com}\\
ORCID: 0009-0000-7054-2014}

\date{}

\begin{document}
\maketitle

\begin{abstract}
Large language models (LLMs) are evaluated for calibration using metrics such as Expected Calibration Error that conflate two distinct components: the model's ability to discriminate correct from incorrect answers (sensitivity) and its tendency toward confident or cautious responding (bias). Signal Detection Theory (SDT) decomposes these components. While SDT-derived metrics such as AUROC are increasingly used, the full parametric framework --- unequal-variance model fitting, criterion estimation, z-ROC analysis --- has not been applied to LLMs as signal detectors. In this pre-registered study, we treat three LLMs as observers performing factual discrimination across 168,000 trials and test whether temperature functions as a criterion shift analogous to payoff manipulations in human psychophysics. Critically, this analogy may break down because temperature changes the generated answer itself, not only the confidence assigned to it. Our results confirm the breakdown with temperature simultaneously increasing sensitivity (AUC) and shifting criterion. All models exhibited unequal-variance evidence distributions (z-ROC slopes 0.52--0.84), with instruct models showing more extreme asymmetry (0.52--0.63) than the base model (0.77--0.87) or human recognition memory (${\sim}0.80$). The SDT decomposition revealed that models occupying distinct positions in sensitivity--bias space could not be distinguished by calibration metrics alone, demonstrating that the full parametric framework provides diagnostic information unavailable from existing metrics.
\end{abstract}


\section{Introduction}

When a large language model (LLM) answers a factual question, two things matter: can the model discriminate correct from incorrect responses, and does its expressed confidence reflect that discrimination? These are fundamentally different problems requiring different interventions. A model that cannot discriminate needs better training data or architectural improvements. A model that discriminates well but expresses confidence poorly needs recalibration, not retraining. Current evaluation practice does not make this distinction.

Expected Calibration Error \citep[ECE;][]{guo2017calibration}, the standard metric, computes the average discrepancy between confidence and accuracy within binned intervals. A model achieves low ECE by being accurate at the rate it claims, regardless of its discriminative ability. ECE says nothing about discrimination per se: a model that is uniformly wrong at a rate matching its confidence achieves perfect calibration with zero sensitivity. The Brier score decomposes into reliability, resolution, and uncertainty, but not into sensitivity and bias. Semantic entropy \citep{kuhn2023semantic} measures uncertainty without decomposing its sources. Confidence elicitation methods \citep{tian2023just} and post-hoc recalibration approaches \citep{shen2024thermometer} improve calibration but likewise operate on the composite metric.

Signal Detection Theory \citep[SDT;][]{green1966signal,macmillan2005detection} provides exactly this decomposition. Developed in psychophysics to characterise perceptual decisions under uncertainty, SDT separates performance into sensitivity ($d'$ or $d_a$) and criterion (c). These parameters are mathematically independent: changing the payoff structure shifts c without affecting $d'$, while degrading stimulus quality reduces $d'$ without shifting c.

SDT-derived metrics are already entering the LLM evaluation literature, though typically in limited forms. \citet{steyvers2025metacognition} review LLM metacognition through the lens of metacognitive sensitivity, using AUROC and the meta-$d'$ framework \citep{maniscalco2012signal,maniscalco2014signal} to quantify how well models' confidence discriminates correct from incorrect answers. \citet{steyvers2025what} demonstrate that implicit confidence measures (token likelihoods) exhibit greater metacognitive sensitivity than verbalised confidence. Earlier, \citet{kadavath2022language} established that LLMs can discriminate questions they answer correctly from those they do not, providing the empirical foundation for treating LLMs as signal detectors. These important advances establish that SDT-adjacent measures reveal meaningful structure in LLM confidence. However, they stop short of applying the full parametric framework: fitting unequal-variance Gaussian SDT models to multi-point ROC data, independently estimating criterion placement, testing specific predictions about how experimental manipulations affect the ROC, and validating with converging forced-choice paradigms. The present work takes this next step.

The analogy between temperature scaling and criterion manipulation motivates a specific, testable prediction. In human SDT experiments, manipulating payoffs shifts the criterion without changing the observer's sensory capacity. Temperature scaling performs a formally analogous operation on the logit distribution, changing the mapping from evidence to response probability without changing the evidence (logits) themselves. If this analogy holds, temperature should shift the LLM's criterion while leaving sensitivity unchanged.

However, there is a critical disanalogy that we flag from the outset: in human experiments, the stimulus is identical regardless of the observer's criterion setting. In LLMs, temperature affects the sampling process, meaning the generated answer, and therefore the trial outcome, can change. The model may follow a different generative path, producing a different answer entirely. This "stimulus dependence on criterion" breaks a core assumption of the payoff-manipulation analogy. Our study was designed to test whether and where this breakdown occurs: does temperature behave as a criterion shift despite the disanalogy, and if not, what is the nature of the departure?

We report a pre-registered study applying the full SDT framework to three LLMs performing factual question-answering across seven temperatures. The study comprises 168,000 confidence-rating trials and 6,000 forced-choice trials across two datasets. All hypotheses, analyses, and preprocessing decisions were registered on the Open Science Framework before data collection.

The study makes three core contributions:

\begin{enumerate}
\item Temperature is not a pure criterion manipulation. It simultaneously changes sensitivity and criterion, a dual effect revealing that the payoff analogy breaks down because the "stimulus" (generated answer) depends on the "criterion" (temperature).
\item LLMs exhibit unequal-variance evidence distributions. z-ROC slopes range from 0.52 to 0.84, with instruct models showing more extreme unequal variance than the base model. This parallels but exceeds findings in human recognition memory (typically $s \approx 0.80$; Mickes et al., 2007).
\item The SDT decomposition reveals structure invisible to calibration metrics. Models with different sensitivity and bias profiles cannot be distinguished by ECE alone, demonstrating the practical value of the decomposition for targeted intervention.
\end{enumerate}

\section{Signal Detection Theory Applied to LLMs}

\subsection{The SDT Framework}

The SDT framework is well established within the psychophysics literature. On each trial, the observer receives evidence drawn from either a noise distribution (signal absent) or a signal distribution (signal present) and applies a criterion to generate a response. Sensitivity ($d'$) is the standardised distance between distribution means; criterion (c) is the threshold location. In the unequal-variance SDT (UVSD) model, the signal distribution has greater variance than the noise distribution, parameterised by the ratio s = $\sigma_{\text{noise}}$/$\sigma_{\text{signal}}$. The z-ROC slope directly estimates s, and z-ROC slopes below 1.0 indicate unequal variance. The sensitivity measure $d_a$ = $\sqrt{2/(1 + s^2)}$ $\times$ (s $\times$ z(HR) -- z(FAR)) accounts for the variance asymmetry \citep{macmillan2005detection}.

Confidence-rating ROC data provide richer information than binary responses: each confidence level acts as a virtual criterion, generating multiple ROC points \citep[Chapter~3]{macmillan2005detection}. The area under the ROC (AUC) provides a non-parametric sensitivity measure requiring no distributional assumptions.

\subsection{The LLM-as-Observer Mapping}

We map SDT onto LLM question-answering as follows. Each factual question is a trial. The model is the observer. The question has a correct answer (signal) or the model produces an incorrect answer (noise). The evidence variable is the normalised log-probability (NLP) of the generated answer: $\text{NLP} = (1/L) \sum \log p(t_i \mid t_{<i})$, where $L$ is the answer length in tokens. Higher NLP indicates the model assigns greater probability to its answer, analogous to stronger sensory evidence.

This mapping is functional, not literal. SDT is applied here as a measurement framework for decomposing performance, not as a cognitive model of LLM processing. Unlike human perception, where the evidence variable is a latent sensory representation, NLP is directly observable. Unlike human signal detection, where the observer passively receives stimuli, the LLM generates the response it then evaluates. SDT has precedent for application beyond perception, including medical diagnosis \citep{swets1996signal}, eyewitness identification \citep{wixted2014signal}, and automated system trust \citep{bartlett2017benchmarking}. In each case, the value of SDT lies not in literal equivalence to sensory detection but in the decomposition it provides.

We note that our multi-point ROC curves are constructed from NLP confidence bins at a fixed temperature, what \citet{maniscalco2014signal} would term "pseudo type-1" ROC curves, built from confidence ratings rather than true criterion manipulations. Temperature variation across conditions does provide something closer to a true criterion manipulation, but it serves as our independent variable rather than the basis for ROC construction. This is methodologically standard in recognition memory, where multi-point ROCs are routinely constructed from confidence ratings \citep{mickes2007direct}.

Why NLP as the evidence variable? NLP captures the model's internal probability assessment of the generated sequence, normalised for answer length. It is the natural analogue of the sensory evidence on which human observers base their responses, the continuous variable underlying the binary correct/incorrect decision. Alternative evidence variables exist (e.g., first-token logit, softmax probability). A pre-registered exploratory analysis comparing NLP with first-token logit evidence was specified (Amendment 2) but is deferred to a companion analysis; we address the implications of this choice below (\S5.2). NLP was chosen because it integrates information across the full answer sequence rather than relying on a single token position.

Temperature is the criterion manipulation. At temperature T, softmax probability is $p_i$ = exp($z_i$/T) / $\Sigma$ exp($z_j$/T). Lower temperatures sharpen the distribution (conservative responding); higher temperatures flatten it (liberal responding). This is formally analogous to payoff manipulation, but with the crucial disanalogy that higher temperatures also change which tokens are sampled, potentially changing the answer itself.

\subsection{What SDT Adds Beyond ROC Analysis}

AUROC is widely computed in the LLM calibration literature \citep{steyvers2025what,geng2024survey,liu2025uncertainty,shorinwa2025survey}. The full SDT framework adds four things:

First, independent criterion estimation. AUROC collapses the ROC to a single number, discarding information about where on the ROC the model operates. The criterion parameter c captures where on the ROC the model operates. Two models with identical AUROC can have very different criteria, and the appropriate intervention differs accordingly.

Second, the unequal-variance model. Fitting the UVSD model reveals whether signal and noise evidence distributions have different variances, a structural property of the model's internal representations that has no analogue in AUROC or ECE. The variance ratio s has direct theoretical interpretation: it reflects how variably the model represents correct versus incorrect answers.

Third, specific, testable predictions. SDT generates precise predictions about how experimental manipulations should affect the ROC. The temperature-criterion hypothesis predicts AUC invariance with criterion shift, a prediction testable by equivalence testing (TOST), not possible with AUROC alone.

Fourth, connection to 70 years of psychophysical theory. The decomposition connects LLM evaluation to a mature theoretical framework with established methodology for diagnosing performance and prescribing interventions.

\subsection{Pre-Registered Hypotheses}

H1 (Temperature-as-Criterion-Shift). For T $\in$ {0.1, 0.3, 0.5, 0.7, 1.0}, AUC remains constant (within simulation-derived equivalence bounds) while c varies monotonically with T.

H2 (Hidden Structure). Models with similar ECE occupy different positions in ($d_a$, c) space.

H3 (Paradigm Convergence). $d_a$ from confidence-rating data converges with $d'_{\text{4AFC}}$ from forced choice across domains.

H4 (Unequal Variance). z-ROC slope < 1.0, indicating $\sigma_{\text{signal}}$ > $\sigma_{\text{noise}}$.

H5 (Domain-Specific Sensitivity). $d_a$ varies across knowledge domains.

H6 (High-Temperature Degradation). At T > 1.0, $d_a$ decreases due to generative-path divergence.

\section{Method}

\subsection{Models}

Three models were evaluated: Llama-3-8B-Instruct \citep{meta2024llama}, Mistral-7B-Instruct-v0.3 \citep{jiang2023mistral}, and Llama-3-8B-Base (no instruction tuning). All were run as Q5\_K\_M quantisations via \texttt{llama-cpp-python} 0.3.16 with Vulkan backend on an AMD RX 7900 GRE (16 GB VRAM). The two instruct models served as primary comparisons; the base model provided an instruction-tuning contrast.

\subsection{Datasets}

TriviaQA \citep{joshi2017triviaqa}: 5,000 questions from the unfiltered set (seed = 42), classified into five domains (History \& Politics: 1,248; Arts \& Literature: 1,167; Geography: 667; Science \& Technology: 634; Unclassified: 1,284) using an LLM-based classifier (pre-registered fallback after Wikipedia API lookup failed; see Supplementary Materials, Deviation 1).

Natural Questions \citep{kwiatkowski2019natural}: 3,000 short-answer questions from NQ-Open, providing a replication dataset.

\subsection{Paradigm A: Confidence-Rating}

Each model generated an answer at each of seven temperatures: T $\in$ {0.1, 0.3, 0.5, 0.7, 1.0, 1.5, 2.0}, yielding 168,000 trials. Each trial recorded the generated answer, NLP, softmax probability, and binary correctness (exact match against verified aliases with difflib.SequenceMatcher $\geq$ 0.85 fallback, plus preamble stripping and refusal detection).

A human spot-check on a 5\% subsample (1,200 judgments) yielded a false-match rate of 1.3\% and a missed-match rate of 30.1\%. The high missed-match rate was driven by verbose generation (32\%), alias incompleteness (57\%), and near-threshold similarity (11\%), not by scoring logic errors. Because missed matches are correct answers scored as incorrect, they deflate hit rates uniformly across conditions, producing conservative $d_a$ estimates (lower bounds on true sensitivity). The relative ordering of conditions is unaffected: the same questions produce the same verbose answers at every temperature. The scoring robustness check across similarity thresholds {0.80, 0.85, 0.90} confirmed max $|\Delta d_a|$ = 0.081.

\subsection{Paradigm B: 4-Alternative Forced Choice}

For 2,000 TriviaQA questions, each model chose among four options (correct answer plus three distractors selected by domain matching, token-length matching, and embedding cosine similarity filtering). Proportion correct was converted to $d'_{\text{4AFC}}$ via the mAFC integral \citep{green1991probability}.

\subsection{Analysis A: Force-Decode}

For each model at T = 1.0, we force-decoded both the correct answer (maximum NLP across TriviaQA aliases) and the model's own incorrect generation, yielding signal and noise NLP distributions from which $d_a$ was computed. Because force-decoding uses fixed token sequences with no sampling, this analysis is temperature-invariant and provides a baseline sensitivity estimate independent of the criterion manipulation. A pre-registered robustness check (Amendment 1) replaced model-generated incorrect answers with randomly sampled incorrect answers to test whether the noise distribution was biased by the model's generation tendencies.

\subsection{SDT Analysis}

ROC curves were constructed from 20 equal-width NLP bins (edges from T = 1.0, held constant across temperatures) with \citet{hautus1995corrections} log-linear correction. UVSD models were fit by maximum likelihood (L-BFGS-B; z-ROC regression initialisation plus 10 perturbation restarts). AUC was computed by trapezoidal integration. Bootstrap 95\% CIs used 10,000 resamples per condition ($\geq$99.95\% convergence in all 42 conditions). Equivalence bounds for H1 were derived from a pre-registered Monte Carlo simulation (10,000 iterations, 20 parameter combinations).

A pre-registered robustness check repeated all analyses with quantile-based (equal-count) bins. Results were stable for Llama-3-Base (mean $|\Delta d_a|$ = 0.011) and Llama-3-Instruct (mean $|\Delta d_a|$ = 0.064) but showed substantial divergence for Mistral (mean $|\Delta d_a|$ = 0.453), attributable to bin sparsity at low temperatures in the equal-width scheme. For Mistral, equal-width $d_a$ estimates should be interpreted with caution; quantile estimates ($d_a$ = 1.45--1.59) are likely more reliable than equal-width estimates ($d_a$ = 1.68--2.15). AUC, being non-parametric, was robust across all models and binning schemes.

\subsection{Pre-Registration and Deviations}

All hypotheses and analysis specifications were registered on OSF before data collection. Seven deviations were documented: (1) domain classification fallback method, (2) Llama-3-Base source repository, (3) NQ dataset source, (4) MLE optimisation specification change, (5) Paradigm B 4AFC implementation, (6) NLL vectorisation, (7) scoring pipeline missed-match rate. Full documentation is provided in Supplementary Materials.

\section{Results}

\begin{table*}[!htbp]
\centering
\caption{SDT parameters at selected temperatures (TriviaQA). Bootstrap 95\% CIs from 10,000 resamples.}
\label{tab:sdt_params}
\scriptsize
\begin{tabular}{@{}llcccc@{}}
\toprule
Model & T & AUC [95\% CI] & $d_a$ [95\% CI] & $c$ [95\% CI] & Slope \\
\midrule
Llama-3-Inst & 0.1 & .738 [.726, .750] & 1.43 [1.32, 1.56] & $-$2.85 [$-$2.88, $-$2.80] & 0.52 \\
 & 0.5 & .748 [.735, .760] & 1.37 [1.27, 1.48] & $-$2.64 [$-$2.73, $-$2.56] & 0.56 \\
 & 1.0 & .773 [.761, .785] & 1.39 [1.31, 1.48] & $-$2.40 [$-$2.49, $-$2.32] & 0.63 \\
 & 1.5 & .795 [.783, .806] & 1.45 [1.37, 1.53] & $-$2.19 [$-$2.28, $-$2.11] & 0.62 \\
 & 2.0 & .809 [.797, .820] & 1.50 [1.42, 1.58] & $-$2.07 [$-$2.16, $-$1.99] & 0.65 \\
\midrule
Mistral-7B-Inst & 0.1 & .620 [.610, .629] & 1.86 [1.51, 4.55]$^\dagger$ & $-$2.88 [$-$2.88, $-$2.88] & 0.56 \\
 & 0.5 & .664 [.654, .674] & 2.03 [1.73, 2.57] & $-$2.85 [$-$2.88, $-$2.80] & 0.52 \\
 & 1.0 & .713 [.702, .724] & 1.97 [1.74, 2.31] & $-$2.69 [$-$2.78, $-$2.61] & 0.57 \\
 & 1.5 & .751 [.740, .762] & 2.10 [1.90, 2.38] & $-$2.67 [$-$2.76, $-$2.58] & 0.55 \\
 & 2.0 & .764 [.753, .775] & 2.15 [1.95, 2.46] & $-$2.56 [$-$2.66, $-$2.46] & 0.57 \\
\midrule
Llama-3-Base & 0.1 & .751 [.738, .764] & 1.03 [0.97, 1.09] & $-$2.51 [$-$2.58, $-$2.46] & 0.77 \\
 & 0.5 & .797 [.785, .808] & 1.24 [1.18, 1.30] & $-$2.23 [$-$2.27, $-$2.19] & 0.77 \\
 & 1.0 & .836 [.825, .846] & 1.45 [1.38, 1.51] & $-$1.92 [$-$1.97, $-$1.86] & 0.78 \\
 & 1.5 & .850 [.839, .860] & 1.51 [1.44, 1.58] & $-$1.62 [$-$1.70, $-$1.55] & 0.79 \\
 & 2.0 & .857 [.847, .867] & 1.57 [1.50, 1.64] & $-$1.43 [$-$1.51, $-$1.37] & 0.79 \\
\bottomrule
\multicolumn{5}{@{}l}{\scriptsize $^\dagger$Wide CI due to bin sparsity at low T; quantile estimate preferred.}
\end{tabular}
\end{table*}

\subsection{H1: Temperature Is Not a Pure Criterion Shift}

\begin{figure*}[htbp]
\centering
\includegraphics[width=\textwidth]{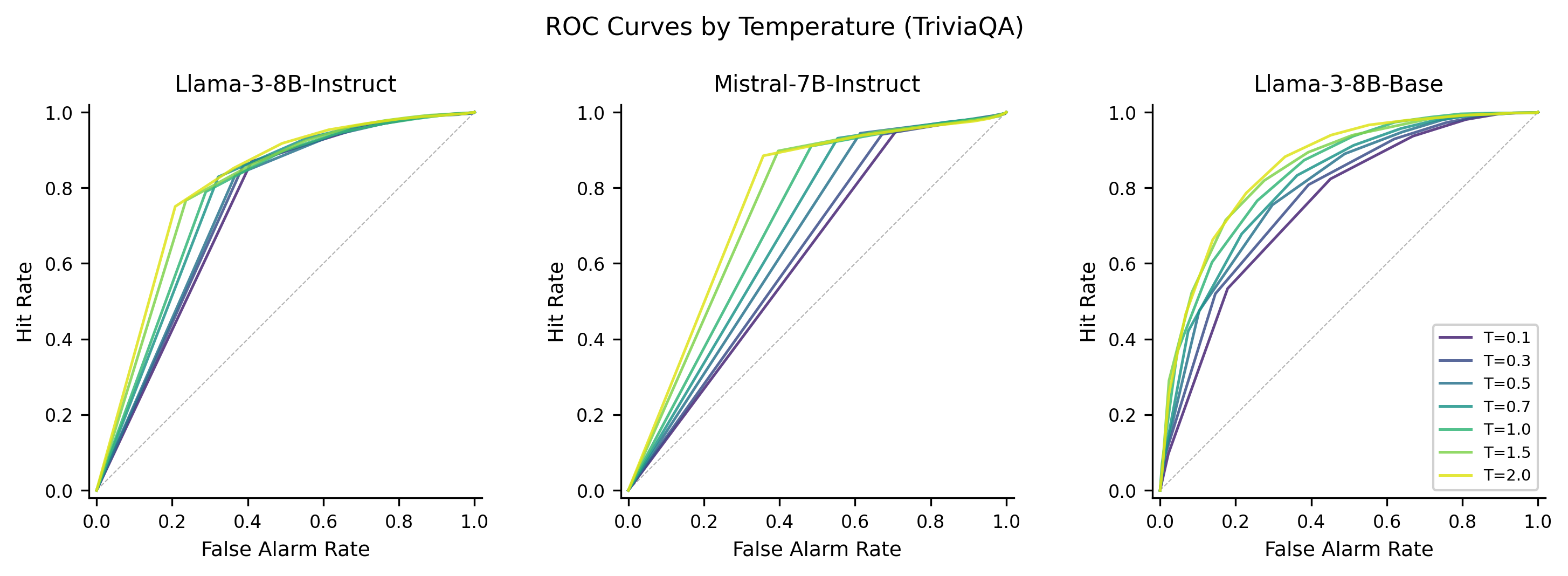}
\caption{ROC curves for three models on TriviaQA across seven temperatures (T = 0.1--2.0). Warmer colours indicate higher temperatures. Curves shift outward with increasing temperature, indicating increasing AUC.}
\label{fig:1}
\end{figure*}

\begin{figure*}[htbp]
\centering
\includegraphics[width=\textwidth]{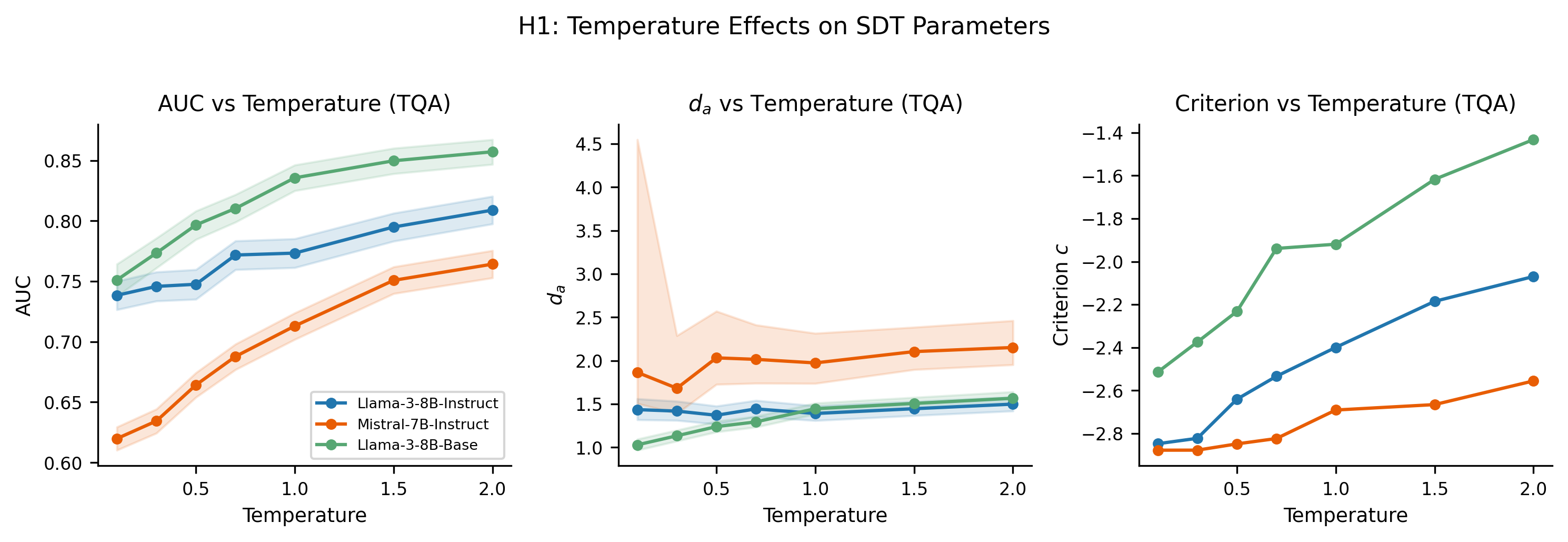}
\caption{Temperature effects on SDT parameters (TriviaQA). Left: AUC increases monotonically ($\rho$ $\geq$ 0.964). Centre: $d_a$ trend. Right: Criterion c shifts monotonically from liberal toward less liberal. Shaded bands: bootstrap 95\% CIs.}
\label{fig:2}
\end{figure*}

The prediction that temperature shifts criterion without affecting sensitivity was not supported. AUC increased monotonically with temperature in all six conditions (Spearman $\rho$ = 1.0 in four conditions, $\rho$ $\geq$ 0.964 in all six; all p < .001). The simulation-derived equivalence bounds (max $\delta$(AUC) = 0.016) were far exceeded: AUC increased by 0.071 to 0.146 from T = 0.1 to T = 2.0.

Simultaneously, criterion shifted as predicted: c increased monotonically with temperature ($\rho$ = 1.0 in five of six conditions), moving from strongly liberal (c $\approx$ --2.88 at T = 0.1) toward less liberal (c $\approx$ --1.40 at T = 2.0).

The rejection of H1 is more informative than confirmation would have been. Temperature simultaneously changes both sensitivity and criterion. Crucially, this sensitivity increase reflects a distributional effect of temperature on the evidence variable: higher temperatures spread the NLP distribution, producing better separation between correct and incorrect answer evidence. It does not indicate improved factual knowledge. Accuracy actually decreased with temperature in all conditions, consistent with \citeapos{renze2024effect} finding that temperature has no significant effect on accuracy in the moderate range (e.g., Llama-3-Base TriviaQA: 59.2\% at T = 0.1, 33.6\% at T = 2.0). AUC is a property of the evidence distribution, not of the model's epistemic competence. The dissociation between AUC (increasing) and accuracy (decreasing) illustrates exactly why the SDT decomposition is needed: aggregate accuracy conflates sensitivity with criterion placement.

\subsection{H2: SDT Reveals Hidden Structure}

\begin{figure*}[htbp]
\centering
\includegraphics[width=\textwidth]{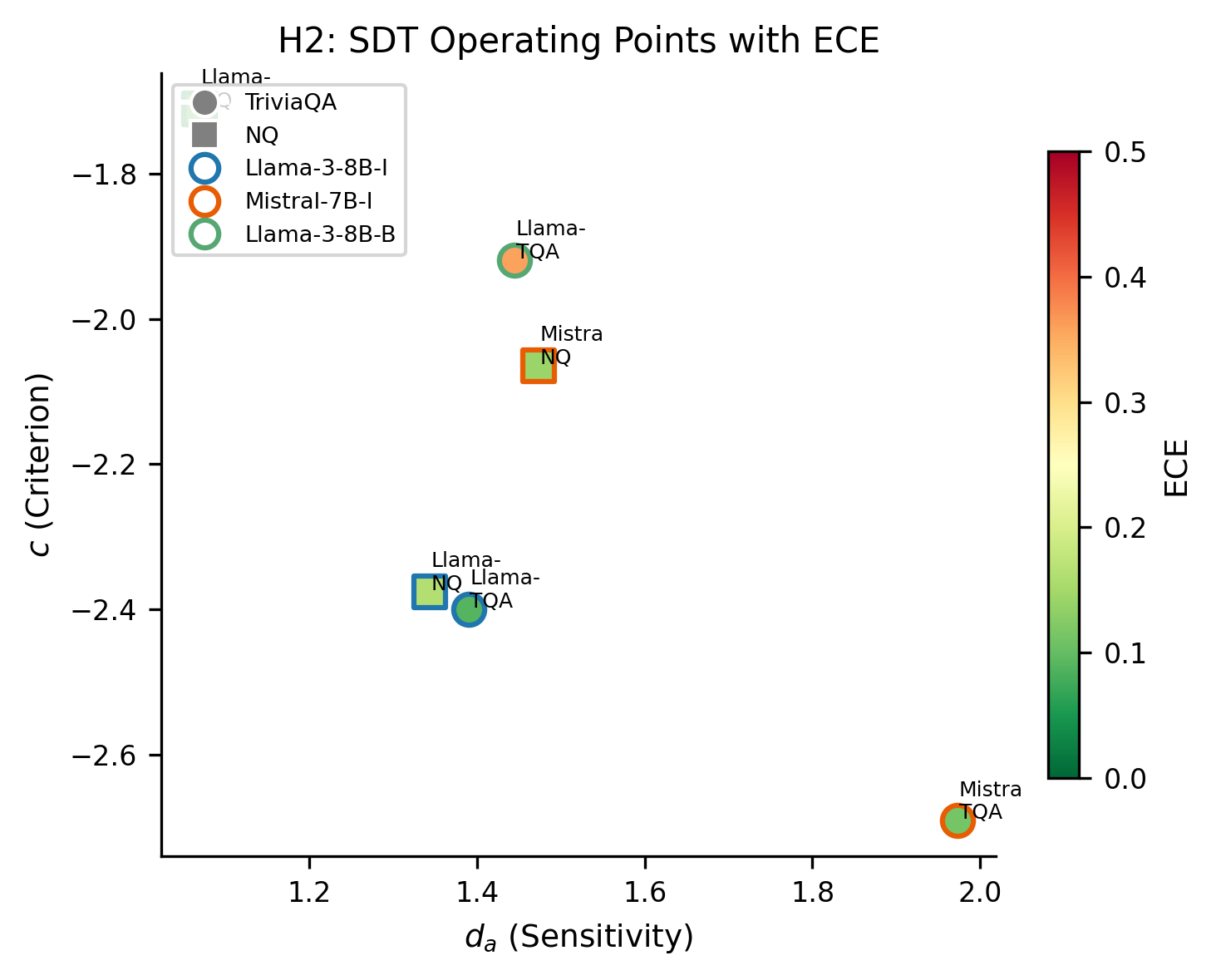}
\caption{SDT operating points at T = 1.0. Position: ($d_a$, c); fill: ECE. Edge colour: model identity. Circles: TriviaQA; squares: NQ. Models occupy distinct positions invisible to ECE alone.}
\label{fig:3}
\end{figure*}

The pre-registered ECE similarity criterion (|$\Delta$ECE| $\leq$ 0.03) was not met (ECE range = 0.27), so H2 was tested under the pre-registered fallback: whether ECE and the SDT operating point provide non-redundant information. At T = 1.0 on TriviaQA, Llama-3-Instruct ($d_a$ = 1.39, c = --2.40, ECE = 0.089) and Mistral ($d_a$ = 1.97, c = --2.69, ECE = 0.112) occupy distinct positions in SDT space: Mistral has higher sensitivity but a more liberal criterion. ECE alone would rank Llama-3 as better calibrated, but SDT reveals that Mistral is the more capable discriminator operating under a more liberal policy. The implied intervention differs: Mistral needs criterion adjustment, not capability improvement.

\subsection{H3: Paradigm Convergence}

\begin{figure*}[htbp]
\centering
\includegraphics[width=\textwidth]{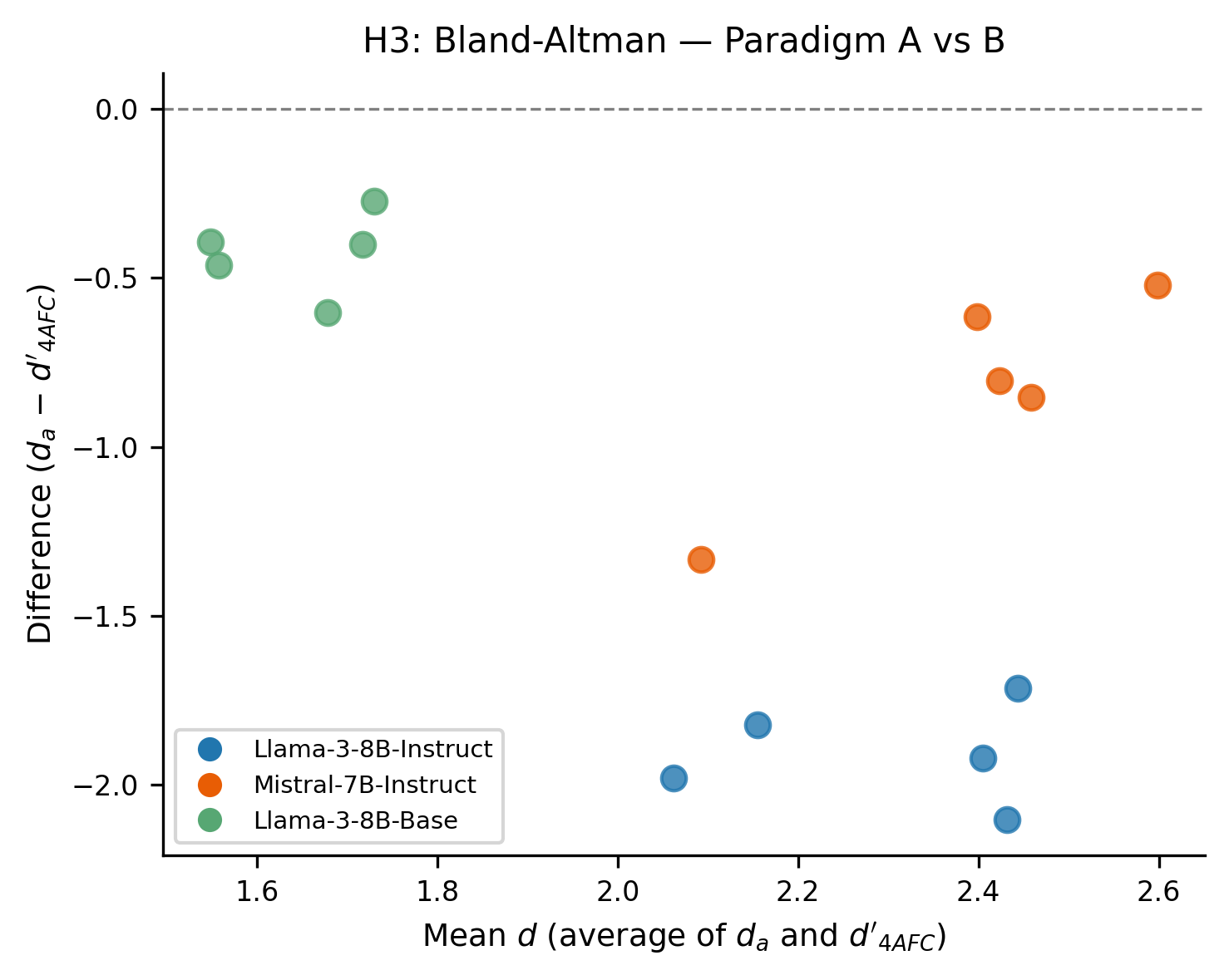}
\caption{Bland-Altman plot: $d_a$ (Paradigm A, T = 1.0) versus $d'_{\text{4AFC}}$ (Paradigm B) across domains. Systematic negative bias reflects near-ceiling 4AFC inflating $d'_{\text{4AFC}}$.}
\label{fig:4}
\end{figure*}

The 4AFC paradigm produced near-ceiling performance for instruct models (Llama-3-Instruct: 97.5\% correct, $d'_{\text{4AFC}}$ = 3.31; Mistral: 94.5\%, $d'_{\text{4AFC}}$ = 2.85; Base: 80.4\%, $d'_{\text{4AFC}}$ = 1.91). Within-model domain correlations between $d_a$ and $d'_{\text{4AFC}}$ were positive but non-significant: Llama-3-Instruct r = .84 (p = .08), Mistral r = .82 (p = .09), Base r = .46 (p = .44), all with N = 5 domains. The non-significance reflects both low power and compression of $d'_{\text{4AFC}}$ variance by ceiling performance: at 97.5\% correct, $d'_{\text{4AFC}}$ is in the steeply nonlinear portion of the psychometric function where small accuracy differences produce large $d'$ changes, squashing between-domain variance. The Bland-Altman plot revealed systematic bias: $d_a$ was consistently lower than $d'_{\text{4AFC}}$ (mean |$\Delta$| = 1.9 $d'$ units). H3 is not supported as specified. The positive correlation direction and preserved domain ordering across paradigms provide qualitative support for the framework, but ceiling effects prevented a quantitative test.

\subsection{H4: Unequal Variance}

\begin{figure*}[htbp]
\centering
\includegraphics[width=\textwidth]{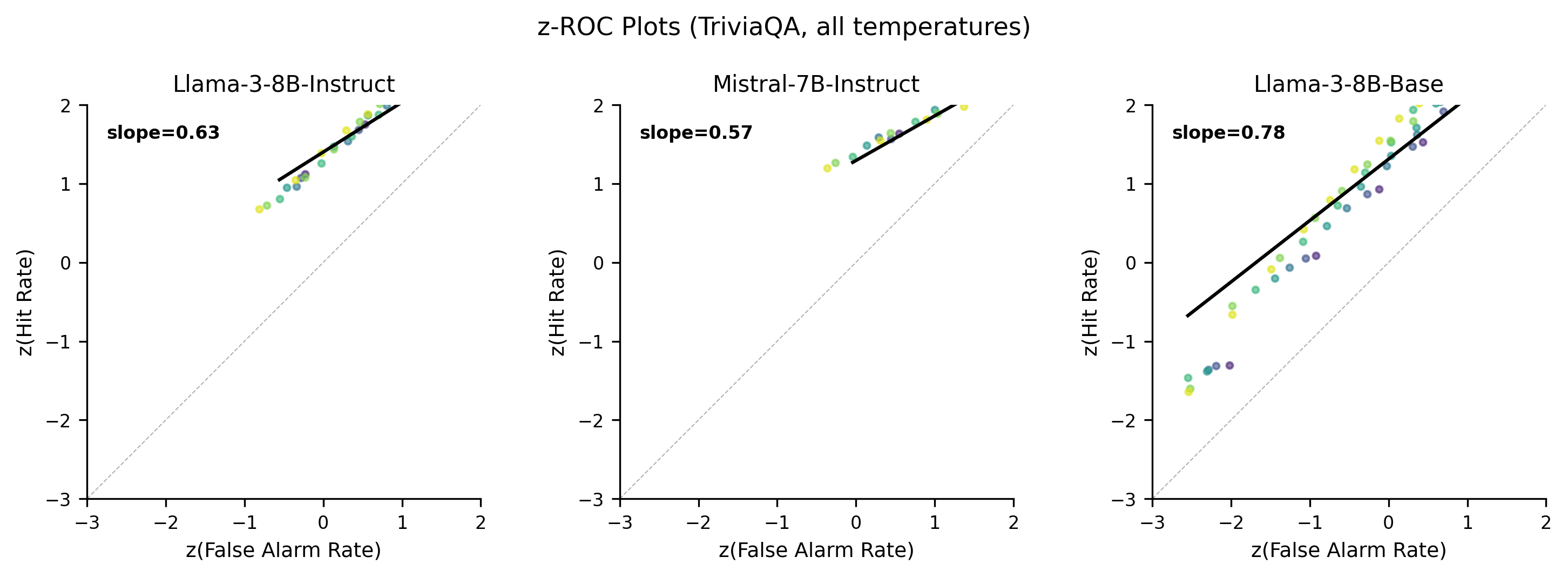}
\caption{z-ROC plots (TriviaQA, all temperatures overlaid). Black line: T = 1.0 regression. Slopes < 1.0 confirm unequal variance. Instruct models: 0.52--0.63; base model: 0.78.}
\label{fig:5}
\end{figure*}

\begin{table}[htbp]
\centering
\caption{z-ROC slopes at T = 1.0 and slope $\times$ temperature correlations (E5). $^*p < .01$.}
\label{tab:zroc}
\small
\begin{tabular}{llccc}
\toprule
Model & Dataset & Slope & $s$ & E5 $\rho$ \\
\midrule
Llama-3-Inst & TQA & 0.63 & 0.88 & 0.93$^*$ \\
 & NQ & 0.58 & 0.75 & 0.89$^*$ \\
Mistral-7B-Inst & TQA & 0.57 & 0.57 & 0.32 \\
 & NQ & 0.54 & 0.68 & $-$0.57 \\
Llama-3-Base & TQA & 0.78 & 1.17 & 0.89$^*$ \\
 & NQ & 0.84 & 1.38 & $-$0.71 \\
\bottomrule
\end{tabular}
\end{table}

z-ROC slope was below 1.0 in all 42 conditions. At T = 1.0 on TriviaQA: Llama-3-Instruct slope = 0.63 (s = 0.88); Mistral slope = 0.57 (s = 0.57); Llama-3-Base slope = 0.78 (s = 1.17). The UVSD model was preferred by AIC and BIC in all conditions.

The instruct models showed more extreme unequal variance than the base model, and more extreme than the modal value in human recognition memory ($s \approx 0.80$; Glanzer et al., 1999; Mickes et al., 2007; Wixted, 2007). A discrepancy between the z-ROC slope and the UVSD-fitted s for instruct models indicates some departure from Gaussianity. The base model showed closer convergence, suggesting more Gaussian-like evidence distributions without instruction tuning.

\subsection{H5: Domain-Specific Sensitivity}

\begin{figure*}[htbp]
\centering
\includegraphics[width=\textwidth]{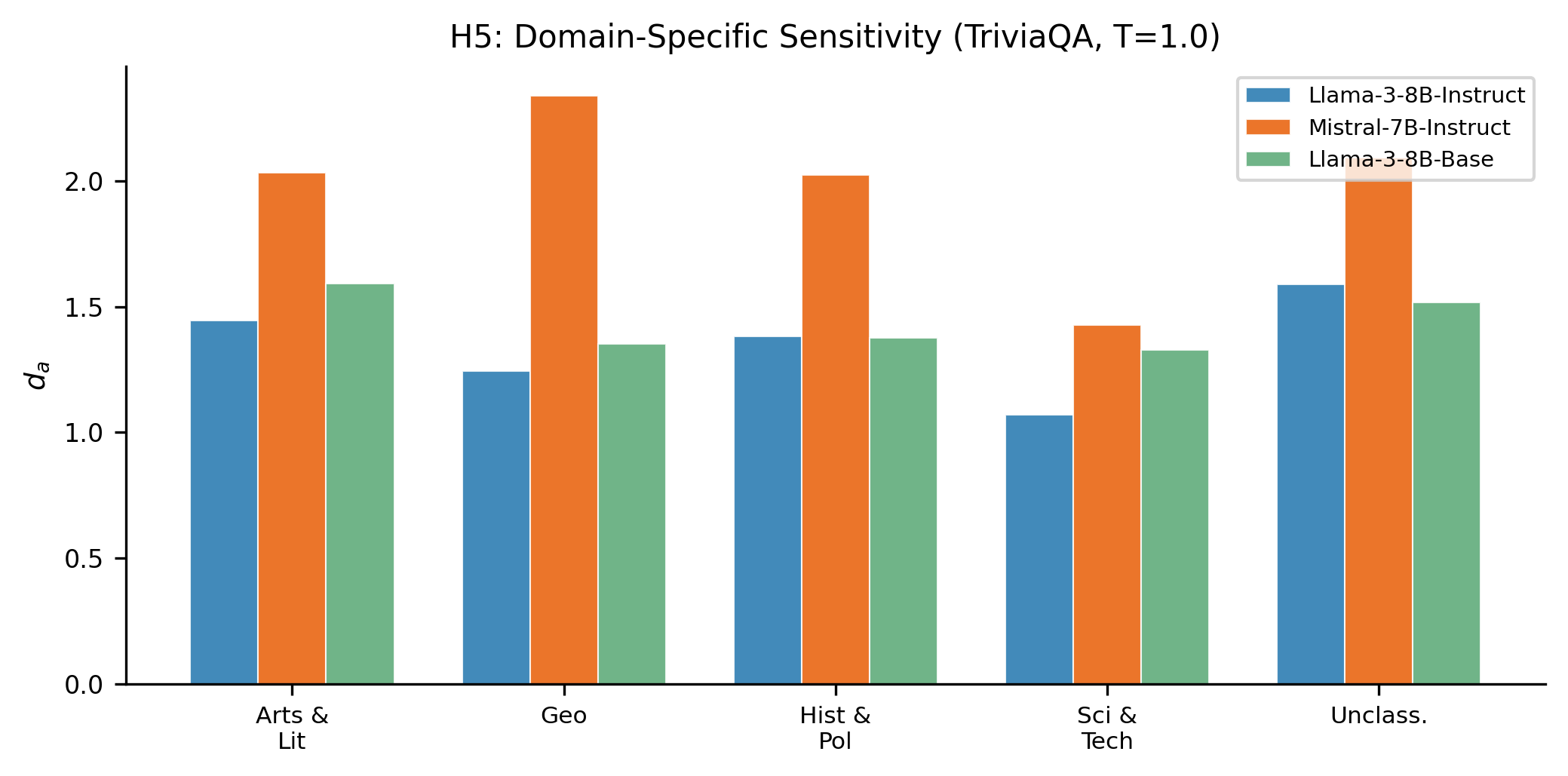}
\caption{Domain-specific $d_a$ at T = 1.0 (TriviaQA). Science \& Technology is consistently hardest. Mistral shows widest range (1.43--2.34).}
\label{fig:6}
\end{figure*}

$d_a$ varied across TriviaQA domains at T = 1.0. Science \& Technology was consistently the hardest domain (Llama-3-Instruct: $d_a$ = 1.07; Mistral: 1.43; Base: 1.33). Mistral showed the widest range (1.43--2.34). This domain-specific information is invisible to aggregate ECE.

\subsection{H6: High-Temperature Sensitivity}

Contrary to predictions, $d_a$ did not decrease at T > 1.0. In five of six conditions, $d_a$ at T = 2.0 exceeded $d_a$ at T = 1.0. For Llama-3-Base on TriviaQA, non-overlapping bootstrap CIs confirmed the increase: $d_a$ = 1.03 [0.97, 1.09] at T = 0.1 versus 1.57 [1.50, 1.64] at T = 2.0. As with H1, this reflects a distributional property of NLP under temperature scaling, not a capability improvement. \citet{kempton2025local} provide a formal account of how temperature distorts local normalisation in autoregressive decoding, consistent with this pattern.

\subsection{Exploratory Analyses}

\begin{figure*}[htbp]
\centering
\includegraphics[width=\textwidth]{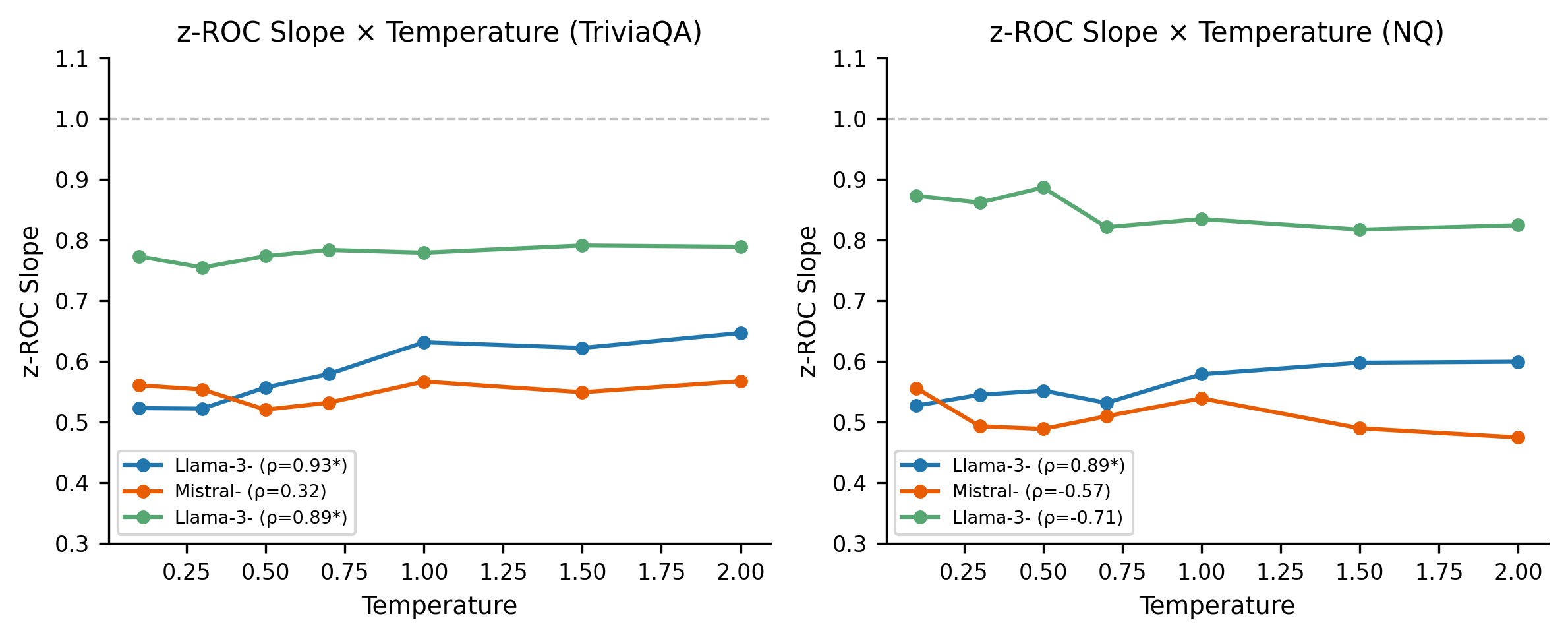}
\caption{z-ROC slope as a function of temperature. Slope increases significantly with T for Llama-3 models ($\rho$ $\geq$ 0.89, p < .01) but not Mistral, indicating model-specific changes in evidence distribution shape.}
\label{fig:7}
\end{figure*}

E1: Instruction tuning as criterion shift. On TriviaQA at T = 1.0, Llama-3-Instruct and Llama-3-Base had similar sensitivity ($\Delta d_a$ = 0.055) but different criteria ($\Delta$c = 0.480), consistent with instruction tuning primarily adjusting response policy rather than discriminative capacity.

E2: Prompt criterion manipulation. Liberal and conservative system prompts did not produce graded criterion shifts. The conservative prompt induced modest refusal (10.6\% for Llama-3-Instruct, 8.0\% for Mistral) alongside small criterion shifts ($\Delta$c = --0.01 to --0.11), but also reduced sensitivity ($\Delta d_a$ = --0.11 to --0.25), suggesting the prompt disrupted processing rather than cleanly shifting criterion.

E5: z-ROC slope $\times$ temperature. z-ROC slope increased significantly with temperature for Llama-3-Instruct ($\rho$ = 0.93, p = .003) and Llama-3-Base ($\rho$ = 0.89, p = .007) on TriviaQA, but not for Mistral ($\rho$ = 0.32, p = .48). Temperature changes the shape of evidence distributions, not just their location, and this effect is model-specific.

\subsection{Robustness}

\begin{figure*}[htbp]
\centering
\includegraphics[width=\textwidth]{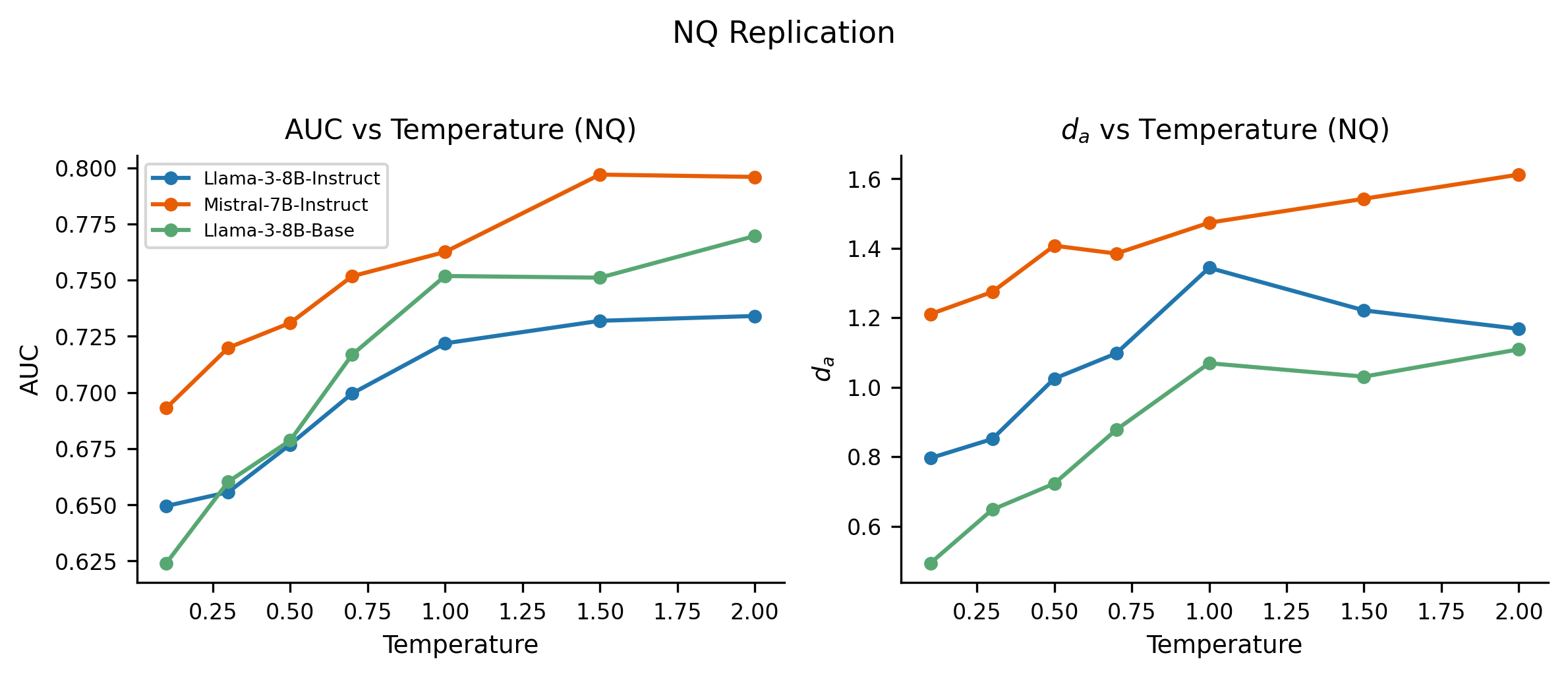}
\caption{NQ replication: AUC and $d_a$ versus temperature. All primary patterns replicate on Natural Questions.}
\label{fig:8}
\end{figure*}

Scoring threshold robustness: max $|\Delta d_a|$ = 0.081, max |$\Delta$AUC| = 0.007 across {0.80, 0.85, 0.90}. Quantile-bin robustness: stable for Llama-3 models, unstable for Mistral (see \S3.6). NQ replication: all primary patterns (AUC increasing with T, criterion shift, unequal variance) replicated on Natural Questions. Evidence variable robustness: a pre-registered analysis comparing first-token logit with sequence NLP as the evidence variable (Amendment 2) was deferred; we address the implications of the NLP fluency confound in \S5.2.

\section{Discussion}

\subsection{Three Core Contributions}

The present study establishes three findings through the application of formal parametric SDT to LLMs as signal detectors.

First, temperature is not a pure criterion manipulation. The payoff analogy from human psychophysics breaks down because temperature changes the generated answer, not only the confidence assigned to it. In human experiments, the stimulus is constant regardless of criterion setting. In LLMs, the evidence variable (NLP of the generated answer) is not independent of the criterion manipulation (temperature) because the answer mediates both. This is a structural disanalogy, not a methodological limitation, and it characterises the boundary of the temperature-as-criterion-shift metaphor.

Importantly, the sensitivity increase observed with temperature is a distributional effect on the evidence variable, not evidence of improved factual competence. Accuracy declined with temperature in all conditions. What increased was the separation of NLP distributions for correct versus incorrect answers, a property of how temperature affects the mapping from logits to probabilities, not a change in what the model knows. This dissociation between AUC (increasing) and accuracy (decreasing) is precisely the kind of structure that the SDT decomposition is designed to reveal.

Second, LLMs exhibit unequal-variance evidence distributions. The z-ROC slopes of 0.52--0.84 consistently indicate that correct-answer evidence has greater variance than incorrect-answer evidence. In human recognition memory, this is explained by encoding variability \citep{ratcliff1992testing,wixted2007dual}: well-remembered items produce high, variable evidence, while unstudied items are more uniform. The LLM analogue is natural: some correct answers are deeply represented in the model's parameters (high NLP) while others are at the edge of the model's knowledge (low NLP, uncertain). Incorrect answers, being the model's best guess, cluster at intermediate NLP values.

The instruct models showed more extreme unequal variance (slopes 0.52--0.63) than the base model (0.77--0.87) and than is typically found in human recognition memory ($s \approx 0.80$). This "hyper-unequal" variance may reflect instruction tuning amplifying the distinction between well-known and edge-of-knowledge items, plausibly because RLHF training teaches the model to express confidence more variably for correct answers.

Third, the SDT decomposition provides information unavailable from calibration metrics. The H2 and H5 results demonstrate that ECE does not distinguish between models that differ in sensitivity versus criterion, nor between domains that differ in discriminability. The E1 result is particularly instructive. Instruction tuning primarily shifts criterion with minimal sensitivity change, a pattern invisible to ECE and directly actionable: it means that if an instruction-tuned model's responses are too liberal, the fix is criterion adjustment (e.g., temperature tuning), not retraining.

\subsection{The Fluency–Accuracy Dissociation}

The force-decode analysis (Analysis A) produced a theoretically informative negative result. Force-decoded correct answers had lower NLP than model-generated incorrect answers in most conditions ($d_a$ $\approx$ 0), yet random incorrect answers (from Amendment 1) produced $d_a$ of 1.7--3.4 against the same correct answers. The model has a genuine truth signal, but it is masked by the high fluency of model-generated errors. The model generates what it finds probable, and its confident incorrect answers are more fluent than externally-sourced correct answers phrased differently from how the model would naturally generate them.

This fluency--accuracy dissociation has direct implications for the primary analyses. Because NLP conflates correctness with fluency, the $d_a$ estimates from Paradigm A should be understood as reflecting the model's ability to discriminate correct from incorrect answers as indexed by their fluency-weighted probability, not by a pure truth signal. The force-decode analysis provides partial validation: when the fluency advantage of self-generated errors is removed (Amendment 1, random noise), $d_a$ rises to 1.7--3.4, confirming that the models do encode a genuine correctness signal that NLP partially captures. The pre-registered first-token logit analysis (Amendment 2), which would construct ROCs from a single-token evidence variable less confounded by fluency, was not completed in the present study and represents an important next step for isolating the truth-tracking component. Despite this limitation, the relative ordering of conditions is unaffected by the fluency confound --- the same fluency bias applies at every temperature and across all models --- so the comparative conclusions (H1 rejection, H2 structure, H4 unequal variance, H5 domain differences) remain valid even if absolute $d_a$ values are conservative.

\subsection{Connections to Metacognitive SDT}

\citet{steyvers2025metacognition} frame LLM confidence evaluation through metacognitive SDT, using AUROC and meta-$d'$ \citep{maniscalco2012signal} to quantify metacognitive sensitivity. Our work complements theirs by moving from descriptive metrics to the full parametric model. The UVSD model fitting, criterion estimation, and z-ROC analysis we report go beyond what AUROC or meta-$d'$ capture: they reveal the variance structure of the evidence distributions, the location of the decision threshold, and how both change under experimental manipulation.

A natural extension would be to compute meta-$d'$ from our data and compare it with $d_a$ from the UVSD fit. If meta-$d'$ < $d_a$, the model's metacognitive sensitivity falls short of what the type-1 evidence supports, indicating metacognitive inefficiency. This connection between parametric SDT and metacognitive SDT was not pre-registered but represents a productive direction for integrating these two approaches to LLM confidence evaluation.

\subsection{Practical Implications}

The decomposition has direct practical value. When a model performs poorly on a domain, ECE does not distinguish "can't discriminate" (low $d_a$) from "discriminates but responds poorly" (wrong c). Our H5 results show Science \& Technology consistently has the lowest $d_a$. This domain needs better training, not recalibration.

The decomposition extends naturally to selective prediction \citep{geifman2017selective}. An abstention threshold is a criterion placement. Given $d_a$ for a specific domain, SDT prescribes the optimal criterion for any desired operating point on the ROC. Current selective prediction methods set thresholds empirically without reference to the underlying sensitivity.

\subsection{Limitations}

Several limitations constrain interpretation. Only two model families (Llama-3-8B and Mistral-7B) were tested, both at 7--8B parameters with Q5\_K\_M quantisation. Whether findings generalise to larger models, different architectures, or full-precision inference is unknown.

The scoring pipeline's 30.1\% missed-match rate, though producing only conservative bias (underestimating sensitivity uniformly across conditions), means that absolute $d_a$ values should not be compared directly with other benchmarks. Relative comparisons within our study remain valid.

The 4AFC paradigm produced near-ceiling performance for instruct models (97.5\% correct), limiting $d'_{\text{4AFC}}$ precision and preventing a clean paradigm convergence test. Future work should use more difficult distractors to bring performance into the 70--85\% range where $d'$ estimation is most precise.

The Mistral $d_a$ estimates were sensitive to the binning scheme, with equal-width bins inflating $d_a$ due to bin sparsity at low temperatures. We report equal-width results as the pre-registered primary analysis but note that quantile estimates are preferred for Mistral.

The NLP evidence variable conflates correctness with fluency (\S5.2). The pre-registered first-token logit comparison (Amendment 2) would partially address this concern but was not completed. The force-decode analysis provides indirect validation that a genuine correctness signal underlies the NLP-based estimates, but a direct comparison of evidence variables remains an important gap. Finally, the evidence distributions show some departure from Gaussianity (z-ROC slope vs. UVSD s discrepancy), particularly for instruct models. This is a known limitation of parametric SDT. The z-ROC linearity test provides a direct diagnostic, and non-parametric AUC provides a robust alternative that requires no distributional assumptions.

\subsection{Future Directions}

Several extensions are immediate. The first-token logit analysis (Amendment 2) is the most pressing: if ROCs constructed from first-token logits produce converging $d_a$ estimates (within the pre-registered $\pm$0.3 bound), this validates the NLP-based results; if they diverge, it would indicate that fluency contamination materially distorts the SDT decomposition. The raw data for this analysis are already collected. Continuous MLE fitting without binning would eliminate bin-sparsity issues. Hierarchical Bayesian SDT models could estimate parameters across conditions simultaneously. Application to larger models and proprietary APIs would test generality. The z-ROC slope $\times$ temperature interaction (E5) warrants mechanistic investigation through analysis of internal representations. And the connection to meta-$d'$ (\S5.3) would formally bridge parametric and metacognitive SDT approaches to LLM evaluation.

More broadly, this work demonstrates that classical psychophysical methods provide genuinely new diagnostic information about AI systems, consistent with the growing programme of applying psychophysics to AI \citep{hernandezcamara2025contrast}. The unequal-variance finding, the domain-specific sensitivity profiles, and the instruction-tuning-as-criterion-shift result are all invisible to standard calibration metrics. As LLMs are deployed in high-stakes domains, the distinction between "can't discriminate" and "discriminates but responds poorly" becomes practically critical. SDT provides the principled framework for making this distinction.

\section*{Declarations}

\paragraph{Funding} The author received no funding for this research. No funds, grants, or other support were received during the preparation of this manuscript.

\paragraph{Competing Interests} The author has no relevant financial or non-financial interests to disclose.

\paragraph{Ethics Approval} Not applicable. This study involved computational experiments with publicly available language models and datasets. No human participants were involved.

\paragraph{Data and Code Availability} All data, analysis code, and raw results are available at \url{https://anonymous.4open.science/r/sdt_calibration-3662/README.md}. The study was pre-registered at \url{https://osf.io/qpk9a/overview?view_only=28f2894d9fcf4679a8afd2b7d70f6f0b}.

\paragraph{Author Contributions} JP Cacioli designed the study, conducted all data collection, performed all analyses, and wrote the manuscript.

\paragraph{Use of Generative AI} During the preparation of this work, the author used Claude (Anthropic) as a research assistant for code generation, statistical analysis scripting, and figure generation. All scientific decisions, hypothesis formulation, and interpretive judgments were made by the author.

\bibliography{references}
\end{document}